%% file: main.tex
\pdfoutput=1

\documentclass[11pt]{article}

\usepackage[]{emnlp2021}
\input{packages}

\input{macros}

\usepackage{hyperref}

\definecolor{mycolor}{RGB}{229, 83, 0}

\newcommand{\OURS}{CLIP-ViL\xspace}
\newcommand{\OURSP}{CLIP-ViL$_\text{p}$\xspace}
\newcommand{\vl}{V\&L\xspace}

\newcommand{\vlpmodel}{CLiP-ViL$_\text{p}$\xspace}

\newcommand{\bertbase}{BERT$_{\textsc{Base}}$\xspace}

\newcommand{\clipres}{CLIP-Res50\xspace}
\newcommand{\clipresbig}{CLIP-Res101\xspace}
\newcommand{\clipresbiggest}{CLIP-Res50x4\xspace}
\newcommand{\clipvit}{CLIP-ViT-B\xspace}

\newcommand{\imagenet}{ImageNet\xspace}
\newcommand{\imagenetres}{\imagenet-Res50\xspace}

\newcommand{\imagenetresbiggest}{\imagenet-Res152\xspace}
\newcommand\blfootnote[1]{%
  \begingroup
  \renewcommand\thefootnote{}\footnote{#1}%
  \addtocounter{footnote}{-1}%
  \endgroup
}

\title{How Much Can CLIP Benefit Vision-and-Language Tasks?}
\author{
Sheng Shen$^{*\dagger}$, Liunian Harold Li$^{*\ddagger}$, Hao Tan$^{\circ}$, 
 Mohit Bansal$^{\circ}$, 
 \\ \textbf{Anna Rohrbach$^{\dagger}$, Kai-Wei Chang$^{\ddagger}$, Zhewei Yao$^{\dagger}$ and Kurt Keutzer$^{\dagger}$}
 \\
 $^\dagger$University of California, Berkeley, $^\ddagger$University of California, Los Angeles\\
 $^\circ$University of North Carolina at Chapel Hill \\
 \texttt{\small \{sheng.s, anna.rohrbach, zheweiy, keutzer\}@berkeley.edu,} 
 \\
 \texttt{\small \{liunian.harold.li, kwchang\}@cs.ucla.edu, \{haotan, mbansal\}@cs.unc.edu}
}

\begin{document}
\maketitle

\input _s0_abstract.tex

\blfootnote{$^*$The two authors contributed equally.}
\input _s1_intro.tex

\input _s2_related_work.tex

\input _s3_method.tex
\input _s4_results.tex
\input _s5_discussion.tex
\input _s6_conclusion.tex

\bibliography{ref}
\bibliographystyle{acl_natbib}

\input _s7_appendix.tex

\end{document}

%% file: packages.tex
\usepackage[utf8]{inputenc} 
\usepackage[T1]{fontenc}    
\usepackage{url}            
\usepackage{booktabs}       
\usepackage{amsfonts}       
\usepackage{nicefrac}       
\usepackage{microtype}      
\usepackage{times}
\usepackage{soul}
\usepackage[utf8]{inputenc}
\usepackage{graphicx}
\usepackage{amsmath}
\usepackage{booktabs}
\usepackage{adjustbox}

\usepackage{graphicx}
\usepackage{subfigure}
\usepackage{booktabs} 

\usepackage{multirow}
\usepackage{paralist}

\usepackage{booktabs} 
\usepackage{multicol}
\usepackage{diagbox}

\usepackage[ruled,noend]{algorithm2e}

\SetCommentSty{mycommfont}

\usepackage{here}

\usepackage{amsmath,amssymb,amsfonts,amsbsy,amsfonts,latexsym}
\usepackage{multirow}
\usepackage{makecell}
\usepackage{xcolor}
\usepackage{colortbl}

\definecolor{colorA}{RGB}{189,201,225}
\definecolor{colorB}{RGB}{103,169,207}
\definecolor{colorC}{RGB}{ 28,144,153}
\definecolor{colorD}{RGB}{  1,108, 89}

\newcolumntype{R}{>{\columncolor{gray!40}}r}
\newcolumntype{L}{>{\columncolor{gray!40}}l}
\newcolumntype{C}{>{\columncolor{gray!40}}c}

\usepackage{tabularx,colortbl,xcolor}
\usepackage{multirow}
\usepackage[normalem]{ulem}
\useunder{\uline}{\ul}{}

\usepackage{enumitem}

\usepackage{xparse}

\DeclareGraphicsExtensions{.pdf,.png}

\SetKwInput{KwInput}{Input}

\usepackage{longtable}

\usepackage{url}
\usepackage{booktabs}
\usepackage{multirow}
\usepackage{array, caption, makecell, booktabs}
\usepackage{wrapfig}
\usepackage{comment}
\usepackage{enumitem}
\usepackage{accents}
\usepackage{amsmath,amsfonts,bm}
\usepackage{pifont}

%% file: macros.tex
\usepackage{amsmath,amsfonts,bm}

\newcommand\fref{Figure~\ref}
\newcommand\tref{Table~\ref}
\newcommand\sref{Section~\ref}

\def\1{\bm{1}}

\DeclareMathAlphabet{\mathsfit}{\encodingdefault}{\sfdefault}{m}{sl}
\SetMathAlphabet{\mathsfit}{bold}{\encodingdefault}{\sfdefault}{bx}{n}



%% file: _s0_abstract.tex
\begin{abstract}

Most existing Vision-and-Language (\vl) models rely on pre-trained visual encoders, using a relatively small set of manually-annotated data (as compared to web-crawled data), to perceive the visual world.
However, it has been observed that large-scale pre-training usually can result in better generalization performance, e.g., CLIP (Contrastive Language-Image Pre-training), trained on a massive amount of image-caption pairs, has shown a strong zero-shot capability on various vision tasks. 
To further study the advantage brought by CLIP, we propose to use CLIP as the visual encoder in various \vl models in two typical scenarios: 1) plugging CLIP into task-specific fine-tuning; 2) combining CLIP with \vl pre-training and transferring to downstream tasks. 
We show that CLIP significantly outperforms widely-used visual encoders trained with in-domain annotated data, such as BottomUp-TopDown. 
We achieve competitive or better results on diverse V\&L tasks, while establishing new state-of-the-art results on Visual Question Answering, Visual Entailment, and \vl Navigation tasks. 

\end{abstract}

%% file: _s1_intro.tex
\section{Introduction}
\label{sec:intro}
Vision-and-Language (V\&L) tasks such as VQA \cite{antol2015vqa} test a system's ability to understand and reason about the semantics of the visual world with the help of natural language. 
Most V\&L models rely on a \textit{visual encoder} to perceive the visual world, which translates the raw pixels into vectors from a representation space. 
Recent works \cite{anderson2018bottom,jiang2020defense,zhang2021vinvl} observe that the visual representation has become the performance bottleneck of V\&L models and stress the importance of learning a powerful visual encoder. 
These high-performing visual encoders are trained on manually-annotated data with class labels (e.g., ImageNet)~\cite{russakovsky2015imagenet} or bounding boxes (e.g., Visual Genome)~\cite{krishna2017visual}. 
However, such detection or image classification data is costly to collect, and the visual representation is limited by the pre-defined class labels.
Thus, there is a need for a visual encoder that is trained on more diverse and large-scale data sources, unbounded by a fixed set of labels, and with generalization ability to unseen objects and concepts.

Recently, CLIP~\cite{radford2021clip} has been proposed to learn visual concepts with language supervision. 
CLIP consists of a visual encoder and a text encoder. It is trained on 400M noisy image-text pairs crawled from the Internet. Thus, the data collection process is scalable and requires little human annotation. 
CLIP has shown strong \textit{zero-shot} capabilities on benchmarks such as ImageNet classification. We hypothesize that it also bears great potential for the V\&L tasks. 
However, directly applying CLIP as a zero-shot model to V\&L tasks 
proves to be difficult (\sref{sec:analysis} and \newcite{kim2021vilt}), 
as many V\&L tasks require complex multi-modal reasoning. 
Thus, we propose to integrate CLIP with existing V\&L models by replacing their \textit{visual encoder} with CLIP's visual encoder.\footnote{Without confusion, we use the term CLIP to interchangeably refer to both the whole CLIP model (including the text and visual encoder) and just its visual encoder. 
}
\begin{figure*}[h!]

    \includegraphics[width=\linewidth]{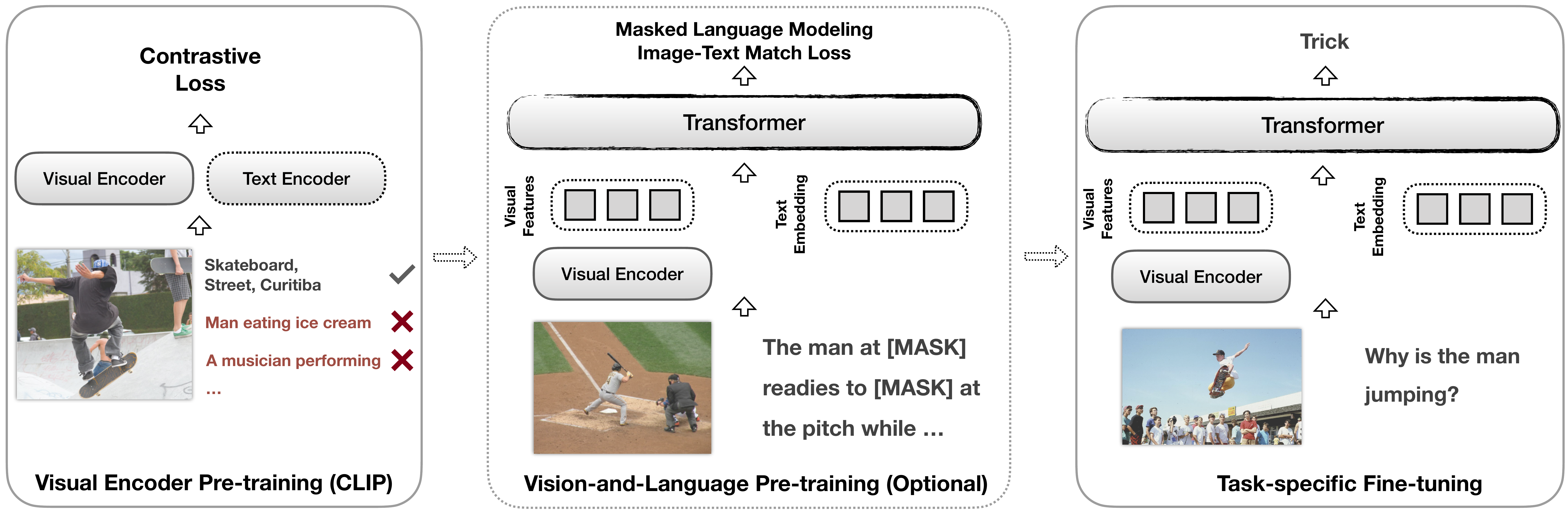}
    \caption{The training process of a V\&L model typically consists of three steps: 1) visual encoder pre-training, 2) vision-and-language pre-training  (optional), and 3) task-specific fine-tuning. In previous V\&L models, visual encoder pre-training requires human annotated vision datasets, which is hard to scale up. Our \OURS proposes to use CLIP, which is trained on image-text pairs crawled from the Internet, as the visual encoder for V\&L models. This reduces the need for human annotated in the pipeline and greatly improves model performance.}
    \label{fig:pipeline}
    \vspace{-10pt}
\end{figure*}

To the best of our knowledge, we present the first large-scale empirical study on using CLIP as the visual encoder for diverse V\&L tasks. 
We consider two typical scenarios: 1) we plug CLIP into direct task-specific fine-tuning (\sref{sec:vl_plug_in}); 2) we integrate CLIP with V\&L pre-training on image-text pairs and transfer to downstream tasks (\sref{sec:vl_pretrain}).\footnote{
We distinguish between \textit{V\&L pre-training} and \textit{CLIP pre-training}:
V\&L pre-training models \cite{lu2019vilbert} have deep interactions between modalities while CLIP follows a shallow-interaction design (\sref{sec:background}).}
For clarity, we denote the models used in these two scenarios as \textbf{\OURS} (without \vl pre-training) and \textbf{\OURSP} (with \vl pre-training).

In \emph{direct task-specific fine-tuning}, we consider three popular tasks: Visual Question Answering~\cite{antol2015vqa}, Image Captioning~\cite{chen2015microsoft}, and Vision-and-Language Navigation~\cite{anderson2018vision}. 
On all three tasks, \OURS brings sizable improvement over strong baselines, 1.4\% accuracy on VQA v2.0, 6.5 CIDEr on COCO Captioning, and 4.0\% success rate on Room-to-Room navigation.

In \emph{V\&L pre-training}, we replace the conventionally used region-based representation~\cite{anderson2018bottom} with CLIP. 
\OURSP performs exceptionally well on three benchmarks, including VQA v2.0, SNLI-VE~\cite{xie2019visual}, and GQA~\cite{hudson2019gqa}, setting a new state-of-the-art (SotA) on VQA (76.70\% on test-std), and SNLI-VE (80.20\% on test).
\OURSP with \clipres outperforms models based on the widely used region-based encoder, BottomUp-TopDown (BUTD) ResNet101~\cite{anderson2018bottom}.  
Moreover, \OURSP with \clipresbiggest surpasses VinVL-ResNeXt152~\cite{zhang2021vinvl}, which is the current SotA and an extreme scale-up attempt with the region-based encoder.

We open-source our code at \url{https://github.com/clip-vil/CLIP-ViL} and hope that our findings inspire future work to explore better visual encoders in \vl tasks.

%% file: _s2_related_work.tex
\begin{figure*}[t]
    \includegraphics[width=\linewidth]{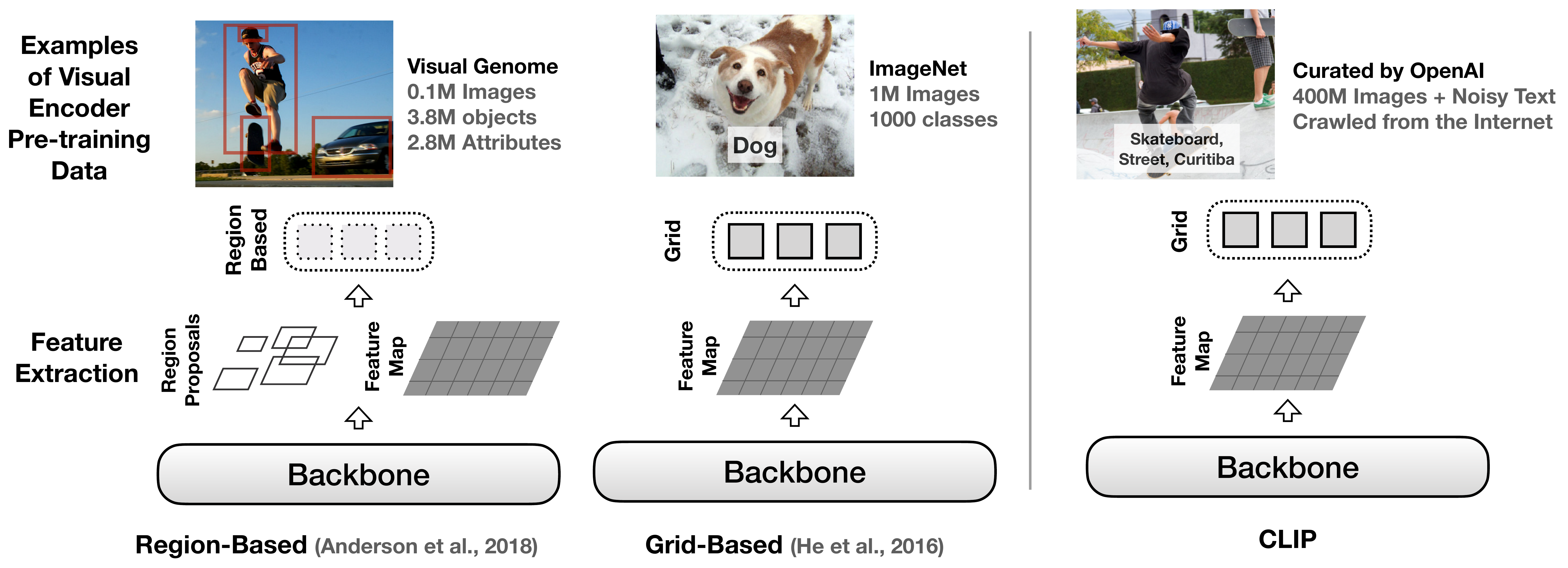}
    \caption{CLIP versus other visual encoders. Region-based methods~\cite{anderson2018bottom} are trained on object detection data. 
    For grid-based methods, previous work use either image classification~\cite{he2016deep} or detection data~\cite{jiang2020defense}. However, CLIP requires only aligned text. }
    \label{fig:visual}
    \vspace{-10pt}
\end{figure*}
\section{Background and Motivation}
\label{sec:background}
\noindent\textbf{Vision-and-Language (V\&L) models.}
V\&L tasks require a model to understand the visual world and to ground natural language to the visual observations. 
Prominent tasks include visual question answering~\cite{antol2015vqa}, image captioning~\cite{chen2015microsoft}, vision-language navigation~\cite{anderson2018bottom}, image-text retrieval~\cite{wang2016learning} and so on. 
V\&L models designed for these tasks often consist of a visual encoder, a text encoder, and a cross-modal interaction module~\cite{kim2021vilt}. 

We illustrate the three typical training stages in~\fref{fig:pipeline}: 
1) the visual encoder is trained on annotated vision datasets~\cite{russakovsky2015imagenet,krishna2017visual} (denoted as \textit{visual encoder pre-training});
2) (optionally) pre-training on paired image-caption data with a reconstructive objective and an image-text matching objective (denoted as \textit{vision-and-language pre-training})~\cite{lu2019vilbert};  
3) fine-tuning on task-specific data (denoted as \textit{task-specific fine-tuning}).

\noindent\textbf{Visual encoders in V\&L models.}
Different models employ different visual encoders, we illustrate their architectures and pre-training processes in \fref{fig:visual}. 
The encoders can be categorized as follows: 1) \textit{region-based} models such as BUTD~\cite{anderson2018bottom} object detector;  2) \textit{grid-based} models such as \newcite{jiang2020defense} that directly extract grid-like feature maps from the visual backbone~\cite{he2016deep, dosovitskiy2020image}.

The encoder is first pre-trained on human-annotated vision datasets.
Region-based encoders are pre-trained with detection data such as Visual Genome~\cite{krishna2017visual}.
Grid-based encoders are pre-trained with image classification data such as ImageNet~\cite{russakovsky2015imagenet} or detection data~\cite{jiang2020defense}. 
However, these manually labeled datasets are expensive to construct and hard to scale up. 
They only provide supervision for a limited number of predetermined visual concepts. 
This motivates us to use CLIP as the visual encoder.

\noindent\textbf{CLIP.} CLIP (Contrastive Language-Image Pre-training) \cite{radford2021clip}\footnote{\href{https://github.com/openai/CLIP}{https://github.com/openai/CLIP}} falls into the line of research that learns visual representations from natural language supervision~\cite{desai2020virtex,sariyildiz2020learning,jia2021scaling}. 
CLIP follows a ``shallow-interaction design'', where a visual encoder and a text encoder encode an input image and text independently, and the dot-product between the two encoder's output is used as the ``alignment score'' between the input image and text. 
It is pre-trained with a contrastive loss where the model needs to distinguish aligned pairs from randomly sampled pairs. 
CLIP leverages an abundantly available source of supervision without human annotation: 400M image-text pairs found across the Internet. 
As a result, CLIP achieves SotA performance in a range of image classification and image-text retrieval tasks in a zero-shot way.

\subsection{Motivation}
Despite the strong zero-shot capability of CLIP on vision tasks, 
CLIP does not exhibit the same level of performance on certain V\&L downstream tasks. 
For instance, if we cast VQA 2.0~\cite{goyal2017making} into a zero-shot image-to-text retrieval task, we only observe chance performance (\sref{sec:analysis}).
Thus, we propose to integrate CLIP's visual encoder with previous V\&L models (\fref{fig:pipeline}). 
We consider the following CLIP variants with different visual backbones \cite{he2016deep,dosovitskiy2020image} (CLIP-ResNet denoted as CLIP-Res): \clipres,  \clipresbig, \clipresbiggest, and \clipvit.
We next describe our methods in two scenarios: 1) direct task-specific fine-tuning (\sref{sec:vl_plug_in})  and 2) V\&L pre-training (\sref{sec:vl_pretrain}).

%% file: _s3_method.tex
\section{\OURS}
\label{sec:vl_plug_in}
In this section, we directly plug CLIP into task-specific models (referred as \OURS) and fine-tune on three representative tasks including Visual Question Answering (\sref{sec:vqa}), Image Captioning (\sref{sec:captioning}), and Vision-Language Navigation (\sref{sec:vln}).

\subsection{Visual Question Answering} 
\label{sec:vqa}
The task of Visual Question Answering (VQA)~\cite{antol2015vqa} is to provide the answer given an image and a related question. Various methods have been introduced~\cite{fukui2016multimodal,yang2016stacked,anderson2018bottom,jiang2018pythia,gao2019dynamic,jiang2020defense}. Here, we select two representative approaches (i.e., Pythia~\cite{jiang2018pythia} and MCAN~\citep{yu2019deep}) to study the impact of the CLIP visual encoders in VQA. 

\noindent\textbf{Experimental Setup.} 
We evaluate on VQA v2.0~\cite{goyal2017making} and follow \cite{jiang2020defense}\footnote{\href{https://github.com/facebookresearch/grid-feats-vqa}{https://github.com/facebookresearch/grid-feats-vqa}} for grid feature extraction. Details of Pythia and MCAN as well as full implementation details are included in the Appendix.

\noindent\textbf{Experimental Results.} 
We report results on the VQA v2.0 Test-dev / Test-std set
in~\tref{tab:vqa_result}. 
Compared to the visual feature extractors pre-trained on \imagenet classification task, CLIP visual modules demonstrate clear improvement (first two blocks of~\tref{tab:vqa_result}). 
\clipres achieves 65.55\%  with Pythia (4.01\% better than \imagenetres) and 71.49\% with MCAN (4.26\% better than \imagenetres) on Test-dev. 
With larger models (i.e., \clipresbig and \clipresbiggest), the results continue improving and the largest \OURS \clipresbiggest outperforms ImageNet-ResNeXt-101$_\text{R}$ (+0.16) with Pythia. 
\OURS \clipresbiggest also achieves the best performance of 74.01\% on Test-dev and 74.17\% on Test-std with MCAN.

We also show results after further detection pre-training on VG~\cite{jiang2020defense}. 
We mark these results as Pythia$_\text{VG}$ and MCAN$_\text{VG}$ in the last two blocks of \tref{tab:vqa_result}.
With \imagenetres encoder, it helps boost the performance by 2.82\% on Pythia$_\text{VG}$ (61.54\% vs. 64.36\%) and 2.90\% on MCAN$_\text{VG}$ (67.23\% vs. 70.13\%). 
However, the performance drops dramatically for \clipres by 5.54\% with Pythia$_\text{VG}$ and 4.08\% with MCAN$_\text{VG}$. 
The potential reason is that CLIP-Res50 is trained on different data and with a different method compared to ImageNet counterparts, so following the previous Visual-Genome fine-tuning practice designed for ImageNet models may hurt. 
We also notice that our best-performing \OURS \clipresbiggest MCAN (74.01\%) still surpasses the best ImageNet-ResNeXt-101 MCAN$_\text{VG}$ (72.59\%) on Test-dev.

\begin{table}[]\resizebox{\linewidth}{!}{
\begin{tabular}{llcc}
\toprule
\multirow{2}{*}{\textbf{VQA Model}}   & \multirow{2}{*}{\textbf{Visual Encoder}} & \multicolumn{2}{c}{\textbf{Result}} \\  \cline{3-4} 
                             &                                  & Test-dev     & Test-std    \\ \hline
\multirow{6}{*}{Pythia}                       & ImageNet-ResNet50*               & 63.21        &       -      \\
                             & \imagenetres                & 61.54        &     61.76        \\
                             & \clipvit                     & 51.38        &      51.47       \\
                             & \clipres                    & 65.55        &     65.78        \\
                             & \clipresbig                    & 66.76        &      67.14       \\
                             & \clipresbiggest                  & \textbf{68.37}        &      \textbf{68.68}      \\\hline
\multirow{5}{*}{MCAN}                         & ImageNet-ResNet50                & 67.23        &      67.46       \\
                             & \clipvit                    & 56.18        &       56.28      \\
                             & \clipres                     & 71.49        &       71.72      \\
                             & \clipresbig                   & 72.77        &       73.19      \\
                             & \clipresbiggest                  & \textbf{74.01}        &        \textbf{74.17}     \\ \hline \midrule
\multirow{5}{*}{Pythia$_\text{VG}$} & \imagenetres *               & 66.27        &          -   \\
    & ImageNet-ResNeXt-101$_\text{R}$*            &  \textbf{68.21}        &  - \\
                             & \imagenet-ResNeXt-101*            & 67.76        &         -    \\
                              & \imagenetres                & 64.36        &      64.81       \\
                             & \clipres                     & 60.01        &       60.06      \\ \hline

\multirow{4}{*}{MCAN$_\text{VG}$}   & ImageNet-ResNeXt-101$_\text{R}$*            & 72.01        &        -     \\
                             & \imagenet-ResNeXt-101*            & \textbf{72.59}        &        -     \\
                              & \imagenetres                 & 70.13        &        70.28     \\
                             & \clipres                     & 67.41        &         67.52   \\ \bottomrule
\end{tabular}
}
\vspace{-8pt}
\caption{Results on VQA v2.0. ``VG'' denotes that the visual encoder has been further pre-trained on Visual Genome detection. ``*'' marks results from \cite{jiang2020defense}. Subscription ``R'' denotes the region features, while other methods use grid features.}
\label{tab:vqa_result}
\vspace{-20pt}
\end{table}

\subsection{Image Captioning}
\label{sec:captioning}
Image captioning aims at generating a natural language description of an image. 
Various methods have been delivered for image captioning \cite{karpathy2015deep,rennie2017self,anderson2018bottom,luo2018discriminability,luo2020better}.
We investigate the effectiveness of the CLIP model for this popular task with \cite{luo2020better} method. 

\noindent\textbf{Experimental Setup.} 
For the model architecture, we experiment with the basic Transformer model adapted from \newcite{vaswani2017attention} in \newcite{luo2020better}.
Grid feature maps are extracted for each image. 
We evaluate our model on COCO dataset~\cite{chen2015microsoft}.  
We use the standard automatic evaluation metrics including CIDEr~\cite{anderson2016spice}, BLEU~\cite{papineni2002bleu}, METEOR~\cite{lavie2007meteor}, and SPICE~\cite{anderson2016spice}. 
The scores are obtained on Karparthy test split~\cite{karpathy2015deep} with beam search of 5 beams. Details are given in Appendix.

\noindent\textbf{Experimental Results.} We report Image Captioning results with different models in~\tref{tab:caption_result}. 
Using the Transformer architecture from \citep{luo2020better}, we see that CLIP-Res models outperform ImageNet pre-trained alternatives for both ResNet50 (+9.1 / +1.5 in CIDEr / SPICE) and ResNet101 (+9.2 / +1.5 in CIDEr / SPICE). 
It even surpasses the strong in-domain region-based feature from BUTD. 
As the model size grows in \OURS, the results also improve and the largest \clipresbiggest achieves the best performance, although there still remains a gap to the pre-trained models that have interactive image-text pre-training phase like Oscar$_\text{base}$ and VinVL$_\text{base}$. 
Again, \clipvit variant leads to dramatically worse performance compared to other visual modules, that we will discuss in \sref{sec:analysis}.

\begin{table}[]\resizebox{\linewidth}{!}{
\begin{tabular}{lcccc}
\toprule
\textbf{Model}             & \textbf{B@4} & \textbf{M} & \textbf{C} & \textbf{S} \\ \hline 
BUTD \cite{anderson2018bottom}                      & 36.3            & 27.7            & 120.1          & 21.4           \\ 
VLP   \cite{zhou2020unified}                     & 39.5            & 29.3            & 129.8          & 22.4           \\ 
AoANet \cite{huang2019attention}                    & 38.9            & 29.2            & 129.8          & 22.4           \\ 
Oscar$_\text{base}$ \cite{li2020oscar}                & 40.5            & 29.7            & 137.6          & 22.8           \\ 
VinVL$_\text{base}$ \cite{zhang2021vinvl}                & \textbf{40.9}            & \textbf{30.9}            & \textbf{140.4}          & \textbf{25.1}           \\  \hline \midrule 
BUTD$_\text{Transformer}$*  \cite{luo2020better}        & -               & -               & \textbf{127.7}          & \textbf{22.5}           \\ 
ImageNet-Res50$_\text{Transformer}$  & 36.2            & 27.6            & 118.8          & 21.2           \\ 
ImageNet-Res101$_\text{Transformer}$ & 36.8            & 27.8            & 121.1          & 21.5           \\ \hline
CLIP-Res50$_\text{Transformer}$      & 38.6            & 28.8            & 127.9          & 22.7           \\ 
CLIP-Res101$_\text{Transformer}$     & 39.2            & 29.1            & 130.3          & 23.0           \\ 
CLIP-Res50x4$_\text{Transformer}$    & \textbf{40.2}            & \textbf{29.7}            & \textbf{134.2}          & \textbf{23.8}           \\ 
CLIP-ViT-B$_\text{Transformer}$     & 21.1            & 19.4            & 58.0           & 12.2           \\ \bottomrule
\end{tabular}}
\vspace{-8pt}
\caption{Image Captioning results. B@4, M, C, and S are BLUE-4, METEOR, CIDEr and SPICE metric, respectively. ``*'' marks results  from~\newcite{luo2020better}. }
\label{tab:caption_result}
\vspace{-5pt}
\end{table}

\subsection{Vision-and-Language Navigation}
\label{sec:vln}
Vision-and-language navigation tests the agent's ability to take action according to human instructions, which recently gains popularity in embodied AI~\cite{anderson2018vision,chen2019touchdown,jain2019stay,chen2019touchdown,qi2020reverie,krantz2020beyond, nguyen2019help, ku2020room}.
Specifically, the agent is put at a location in the environment~\cite{chang2017matterport3d} and asked to reach a target by following the language instructions.
Here, we investigate the impact of the CLIP visual encoder on this new task.

\noindent\textbf{Model Architecture.}
We experiment with the basic attentive neural agent as in~\newcite{fried2018speaker} (please refer to the original paper for implementation details).
At each time step, the agent attends to the panoramic views and the instruction to make an action. 
We replace the pre-trained visual encoder from ImageNet pre-trained ResNet to the pre-trained CLIP visual encoders.
Different from the VQA task that uses a feature map to include detailed information, we use a single-vector output for the entire image following previous works~\cite{fried2018speaker}.
For \clipvit models, we take the output of the [CLS] token. 
For CLIP-ResNet models, we take the attentive pooled feature~\cite{radford2021clip} of the feature map. 
These features are also linearly projected and L2-normalized as in the CLIP model.

\begin{table}[t] \small
\newcommand{\tabincell}[2]{\begin{tabular}{@{}#1@{}}#2\end{tabular}}
\begin{center}
\begin{tabular}{lcc}
\toprule
\multirow{2}{*}{\bf Method} & \multicolumn{2}{c}{\bf Unseen Test} \\
\cmidrule(lr){2-3}
 & \multicolumn{1}{c}{\bf SR}  &\multicolumn{1}{c}{\bf SPL} 
\\ 
\midrule
\multicolumn{3}{l}{\footnotesize{\textit{No Pre-Training}}}\\
                R2R~\cite{anderson2018vision}  
                                        & 20    &  18    \\
                RPA~\cite{wang2018look}    
                                        & 25    & 23  \\
                S-Follower~\cite{fried2018speaker}
                                        & 35    & 28   \\
                RCM~\cite{wang2019reinforced}
                                        & 43    & 38    \\
                SMNA~\cite{ma2019self}  
                                        & 48    & 35      \\
                Regretful~\cite{ma2019regretful}
                                        & 48    & 40    \\
                FAST-Short~\cite{ke2019tactical}
                                        & 54    & 41    \\
                EnvDrop~\cite{tan2019learning}    
                                        & 51    & 47    \\
                PRESS~\cite{li2019robust}
                                        & 49    & 45    \\
                ALTR~\cite{huang2019transferable}
                                        & 48    & 45  \\
                CG~\cite{anderson2019chasing}
                                        & 33    & 30      \\
                RelGraph~\cite{hong2020language}
                                        & 55    & 52    \\
                \textbf{EnvDrop + \OURS} & \textbf{59}    & \textbf{53}     \\
\midrule
\multicolumn{3}{l}{\footnotesize{\color{gray} \textit{Pre-Training}}}\\
                {\color{gray}AuxRN~\cite{zhu2020vision}}
                                        & \color{gray} 55    & \color{gray} 50    \\
                \color{gray} PREVALENT~\cite{hao2020towards}
                                        & \color{gray} 54    & \color{gray} 51    \\
                \color{gray} VLN-BERT\cite{hong2021vlnoe}$\mbox{+}$OSCAR
                                        & \color{gray} 57    & \color{gray} 53    \\
                \color{gray} VLN-BERT\cite{hong2021vlnoe}
                                        & \color{gray} 63    & \color{gray} 57    \\
\bottomrule
\end{tabular}
\vspace{-5pt}
\caption{Unseen test results for Room-to-Room (R2R) dataset. `SR' and `SPL' are Success Rate and Success rate normalized by Path Length. `Pre-Training' methods are mostly in-domain pre-trained on the Matterport3D~\cite{chang2017matterport3d} environments.}

\label{tab:r2r_test}
\vspace{-10pt}
\end{center}
\end{table}

\begin{table}[t] \small
\newcommand{\tabincell}[2]{\begin{tabular}{@{}#1@{}}#2\end{tabular}}
\begin{center}
\begin{tabular}{lcc}
\toprule
\multirow{2}{*}{\bf Method} & \multicolumn{2}{c}{\bf Unseen Test} \\
\cmidrule(lr){2-3}
 & \multicolumn{1}{c}{\bf SR}  &\multicolumn{1}{c}{\bf nDTW} 
\\ 
\midrule
                Random-Baseline~\cite{ku2020room} 
                                        & 7.5    & 15.4    \\
                Mono-Baseline~\cite{ku2020room}
                                        & 25.4    & 41.1  \\
                SAA~\cite{li2021improving}
                                        & 35.4    & 46.8  \\
                \textbf{EnvDrop + \OURS} 
                                        & \textbf{38.3}    & \textbf{51.1}     \\
\bottomrule
\end{tabular}
\vspace{-5pt}
\caption{Unseen test results for Room-across-Room (RxR) dataset under mono-lingual setup. `SR' and `nDTW' are Success Rate and normalized Dynamic Time Warping.
}
\label{tab:rxr_test}
\vspace{-20pt}
\end{center}
\end{table}

\begin{table*}[]
\centering
\resizebox{\textwidth}{!}{%
\begin{tabular}{@{}lcccccccccccc@{}}
\toprule
\multicolumn{1}{c}{\multirow{1}{*}{Features}}  & \multicolumn{4}{c}{Room-to-Room}                         & \multicolumn{8}{c}{Room-across-Room}                                                 \\ 
\cmidrule(l){2-5} \cmidrule(l){6-13} 
\multicolumn{1}{c}{}                          & \multicolumn{2}{c}{Agent} & \multicolumn{2}{c}{BT-Agent} & \multicolumn{2}{c}{English} & \multicolumn{2}{c}{Hindi} & \multicolumn{2}{c}{Telugu} & \multicolumn{2}{c}{Average} \\
\multicolumn{1}{c}{}                          & SR      & \emph{SPL}    & SR        & \emph{SPL}     & SR       & \emph{nDTW}    & SR      & \emph{nDTW}   & SR      & \emph{nDTW}   & SR      & \emph{nDTW}   \\ \midrule
\imagenetresbiggest                               & 48.2    & 44.4            & 53.5      & 48.8             & 35.3     & 50.6             & 37.9    & 51.9            & 37.1    & 52.0  & 36.8  & 51.5          \\
\clipres                     & 52.6    & 47.4            & 56.2      & 49.7             & 38.8     & 53.3             & 44.1    & 55.7            & 43.5    & 55.5  & 42.1     &54.8   \\
\clipvit                                                  & 52.5    & {47.7}            & {57.4}      & {51.3}             & 40.2     & 52.5             & 44.3    & 55.0            & 42.1    & 54.6  & 42.2 & 54.0            \\
\clipresbig                                               & {53.6}    & 47.5            & 56.7              & 49.5    & \textbf{41.0}      & 54.6             & \textbf{44.9}   &\textbf{56.9}            & 42.2    & 55.3 & \textbf{42.7} & 55.6            \\
\clipresbiggest                                                & \textbf{54.7}   & \textbf{48.7}          & \textbf{59.2}      &\textbf{52.9}            &  40.8    & \textbf{54.7}             & 44.5    & 56.5            & \textbf{42.4}    &  \textbf{56.0}   & 42.6 & \textbf{55.7}         \\ \bottomrule
\end{tabular}%
}
\vspace{-8pt}
\caption{Results of Room-to-Room (R2R) and Room-across-Room (RxR) datasets with original ResNet features and CLIP feature variants. 
`BT-Agent'  is the agent trained with back translation (BT). 
`SR' is Success Rate. `SPL' and `nDTW' are the main metrics for R2R and RxR, respectively.  
The best results are bold. \OURS shows clear improvements over the previous ImageNet-trained ResNet model. } 
\label{tab:nav_ablation}
\vspace{-15pt}
\end{table*}

\noindent\textbf{Experimental Setup.}
We apply our model to two vision-and-language navigation datasets: Room-to-Room (R2R, \newcite{anderson2018vision}) and Room-across-Room (RxR, \newcite{ku2020room}).
R2R is built on the indoor environments from the MatterPort3D dataset~\cite{chang2017matterport3d}.
The environments are split into training, unseen validation, and unseen test.
RxR extends the R2R dataset to multiple languages and follows the environment split.
For R2R dataset, we follow the hyperparameter of the publicly available implementation\footnote{\href{https://github.com/airsplay/R2R-EnvDrop}{https://github.com/airsplay/R2R-EnvDrop}} R2R-EnvDrop~\cite{tan2019learning} and replace the input features\footnote{\href{https://github.com/peteanderson80/Matterport3DSimulator}{https://github.com/peteanderson80/Matterport3DSimulator}} with the CLIP features.
For RxR dataset, we change the path length and instruction length; details are given in Appendix.

\noindent\textbf{Experimental Results.}
We show the test-unseen results of our best model (\clipresbiggest) and the comparison to the previous methods.
On R2R dataset (in~\tref{tab:r2r_test}), \OURS reaches 8\% higher in SR (success rate) and 6\% higher in SPL (Success Rate normalized by Path Length) than our baseline, EnvDrop.
\OURS outperforms previous non-pre-training agents and shows competitive results to VLN-specific pre-trained models.
On RxR dataset (\tref{tab:rxr_test}), \OURS achieves the best success rate and nDTW (normalized Dynamic Time Warping) under the mono-lingual setup~\cite{ku2020room} and is 4.3\% better then the previous results for nDTW.

In~\tref{tab:nav_ablation}, we compare different CLIP variants with the previous standard ResNet-152 feature extractors.
These extractors are pre-trained on ImageNet and use the mean-pooled features as the representation for the image.
\clipres shows a clear improvement over the IN alternative (`ImageNet-Res152').
With larger models (i.e., `\clipresbig' and `\clipresbiggest'), the agent performance scales well on both R2R and RxR.
Lastly, we find that the CLIP ViT model (`\clipvit') has similar results as \clipres model. 
ViT also shows a relatively better result when back translation (BT) is applied.
The success of ViT model in VLN is possibly due to the use of [CLS] feature instead of the feature map.

\begin{table*}[h]
\begin{center}
\resizebox{\linewidth}{!}{
\begin{tabular}{l@{\hskip9pt} | 
c@{\hskip9pt}|c@{\hskip9pt}
c@{\hskip9pt}|c@{\hskip9pt}c@{\hskip9pt}
c@{\hskip9pt}c@{\hskip9pt}c@{\hskip9pt}c@{\hskip9pt}
c@{\hskip9pt}c@{\hskip9pt}c@{\hskip9pt}}
\toprule

\multirow{2}{*}{Model} & \multirow{2}{*}{VisualEncoder}  & \multicolumn{2}{c|}{V\&L Pretrain} & & 
\multicolumn{2}{c}{VQA} & & \multicolumn{2}{c}{SNLI-VE} & & \multicolumn{2}{c}{GQA} \\

 &  & Data & Epoch & &Test-Dev & Test-Std  & & Dev & Test-P & & Test-Dev & Test-Std\\
\midrule

PixelBERT  & ImageNet-Res50  & 5.5M & 40 & & 71.35 & 71.42 & & - & - & & - & -\\
PixelBERT  & ImageNet-ResX152  & 5.5M & 40 & & 74.45 & 74.55  & & - & - & & - & -\\
\midrule

LXMERT  & BUTD-Res101  & 9.2M & 20 & & 72.42 & 72.54 & & - & - & & 60.00 & 60.30\\
UNITER  & BUTD-Res101  & 6.5M & - & & 72.70 & 72.91 & & 78.59 & 78.28 & &  - & - \\
Oscar  & BUTD-Res101  & 6.5M & 118 & & 73.16 & 73.44 & & - & - & & 61.19 & 61.23 \\
\midrule
VinVL  & VinVL-ResX152  & 8.9M & 116 & & 75.95 & 76.12 & & - & - & &  \textbf{65.05} & \textbf{65.65} \\

\midrule
\multirow{2}{*}{\textbf{\vlpmodel}} 
& CLIP-Res50  & 9.2M & 20 &
& 73.92 & 74.09 & & 78.64 & 78.97 & & 59.79 & 60.55  \\
& CLIP-Res50x4  & 9.2M & 20 & 
& \textbf{76.48} & \textbf{76.70} &  & \textbf{80.61} & \textbf{80.20} & & 61.42 & 62.93 \\

\bottomrule
\end{tabular}
}
\caption{Evaluation results on three vision-and-language tasks. Our model with \clipres outperforms most BUTD-based models. 
Our model with \clipresbiggest sets a new state-of-the-art on VQA and SNLI-VE. It surpasses VinVL, which is a scaled-up version of BUTD and undergoes more intensive V\&L pre-training than ours.
}
\label{table:vlp_main}
\end{center}
\vspace{-15pt}
\end{table*}

\section{Vision-and-Language Pre-training}
\label{sec:vl_pretrain}
Recently, V\&L pre-training has been proposed as an effective technique to improve the performance on various V\&L tasks~\cite{lu2019vilbert,tan2019lxmert,li2019visualbert,su2019vl,chen2020uniter,zhou2020unified,huang2020pixel,li2020oscar,zhang2021vinvl,li2021unsupervised}. 
Before task-specific fine-tuning, the model is pre-trained on aligned image-text data with a reconstructive objective and an image-text matching objective.
We seek to test the potential of combining CLIP pre-training and V\&L pre-training.
We introduce \vlpmodel, a vision-and-language model pre-trained on image-text data with CLIP visual encoder as its visual backbone. 
In the following, we introduce the model architecture and pre-training process of \vlpmodel in detail.

\subsection{\vlpmodel}
\noindent\textbf{Model Architecture.}
\vlpmodel assumes a text segment $T$ and an image $I$ as input. As in BERT, the text is tokenized into a sequence of subwords $\{w_1, w_2, ..., w_k\}$. 
Every subword is embedded as the sum of its token, position, and segment embeddings \cite{devlin2018bert} and thus the text is embedded as a sequence of word embeddings $\{\bm{w_1}, \bm{w_2}, ..., \bm{w_n}\}$.
The image is embedded as a set of visual vectors $\{\bm{v_1}, \bm{v_2}, ..., \bm{v_m}\}$ from the grid-like feature map.  
The text and visual input are then concatanated into a sequence, 
$\{\bm{w_1}, \bm{w_2}, ..., \bm{w_n}, \bm{v_1}, \bm{v_2}, ..., \bm{v_m}\}$, and processed by a single Transformer. 
In most region-based models, the visual backbone is frozen as fine-tuning the object detector along with the Transformer remains an open problem \cite{su2019vl}. 
In \vlpmodel, the CLIP backbone is trained during both V\&L pre-training and task-specific fine-tuning (see discussion in \sref{sec:analysis}).

\noindent\textbf{Pre-training on Image-Text Data.} 
To learn unified representations for both vision and language, we follow prior work and pre-train the model on image-text pairs. 
We consider three pre-training objectives from LXMERT~\cite{tan2019lxmert}: 1) grounded masked language modeling, where we randomly mask out 15\% of words in the input sentence and train the model to reconstruct the masked words; 2) text-image matching, where the model is provided with a mismatched sentence with a probability of 0.5, and is trained to classify whether the text corresponds to the image; 3) visual question answering, where we train the model to predict the correct answer given a question.

\subsection{Experiments}
\noindent\textbf{Setup.}
We experiment with two variants of CLIP as the visual encoder, \clipres and \clipresbiggest. 
Following LXMERT, we use the same corpora aggregated from MS~COCO Captions~\cite{chen2015microsoft}, Visual Genome Captions~\cite{krishna2017visual}, VQA~\cite{antol2015vqa}, GQA~\cite{hudson2019gqa}, and VG-QA~\cite{zhu2016visual7w} for pre-training. 
We follow the same pre-processing procedure and exclude any test data from the pre-training dataset. This results in 9.18M image-text pairs.

For computational efficiency, we use a relatively small resolution for images. 
We resize the shorter edges of images to 384 and the longer edges to under 640 with preserved aspect ratios. 
During pre-training, as the number of image patches is large, we randomly sample 100 image patches for every image following PixelBERT~\cite{huang2020pixel}. 
We pre-train the model for 20 epochs and unfreeze the CLIP backbone during pre-training and fine-tuning. For details see the Appendix.

\noindent\textbf{Tasks.}
For evaluation, we fine-tune the pre-trained model on three V\&L tasks: VQA v2.0~\cite{goyal2017making}, visual entailment SNLI-VE~\cite{xie2019visual}, and GQA~\cite{hudson2019gqa}. We provide more details in the Appendix. 

\noindent\textbf{Results.}
We report the results in~\tref{table:vlp_main}. We include previous best pre-trained V\&L models and their V\&L pre-training data and epochs. 
As our model is based on \bertbase, we compare only with models based on \bertbase. 
The models are grouped by their visual encoder type. 
We first note that our two models perform competitively on all metrics. 
Especially, \OURS with \clipresbiggest establishes a new SotA on VQA and SNLI-VE.

When comparing with the BUTD visual encoder trained on \textit{in-domain data} (including LXMERT~\cite{tan2019lxmert}, UNITER~\cite{chen2020uniter}, and Oscar~\cite{li2020oscar}), our two models (\OURS with \clipres and \clipresbiggest) significantly outperform most BUTD-Res101 based models. We especially note that LXMERT is trained on the same pre-training dataset and for the same number of epochs as our model, yet our \vlpmodel with \clipres outperforms LXMERT on VQA by 2.59.

VinVL~\cite{li2020oscar} is an extreme scale-up of the region-based paradigm, which is pre-trained on multiple object detection datasets, including MS~COCO~\cite{lin2014microsoft}, OpenImages \cite{kuznetsova2020open}, Object365 \cite{shao2019objects365}, and Visual Genome~\cite{krishna2017visual}. Yet, our model with \clipresbiggest outperforms VinVL on VQA, while requiring significantly less steps of V\&L pre-training. On GQA, our model under-performs VinVL. The potential reason is that GQA is automatically constructed from object bounding box data, which may give region-based models trained on such object data a significant advantage.

Lastly, we compare to Pixel-BERT~\cite{huang2020pixel}, which takes a similar design as our model, but with an ImageNet initialized ResNet. CLIP initialization clearly holds advantage over ImageNet initialization, as \clipres significantly outperforms Pixel-BERT with \imagenetres. 

%% file: _s4_results.tex
\section{Analysis}
\label{sec:analysis}
In this section, we provide detailed analyses on a few interesting phenomena we observe during our experiments, which may help guide future exploration.
Quantitative and qualitative analysis are provided to support our findings.

\noindent\textbf{Zero-Shot Performance of CLIP in VQA.}
In the original paper, CLIP is intended as a zero-shot model and shows strong performance on various vision and image retrieval tasks. 
We are thus curious if CLIP can also perform well as a zero-shot model on \vl tasks that may require complex reasoning.
To conduct zero-shot image classification, CLIP~\cite{radford2021clip} uses the names of all classes in the dataset as the set of candidate text and predict the most probable (image, text) pair.
We thus experiment with a similar setting on VQA but modify the candidate text to be the concatenation of question and answer pair for each question. 
Moreover, \citet{radford2021clip} find a result improvement from prompt engineering.
We follow this design by constructing ``question: [question text] answer: [answer text]'' as the prompt template. 
The results on VQA v2.0 \texttt{mini-eval} are shown in \tref{tab:zero-shot_vqa}.  
All CLIP variants perform at near-chance level in the zero-shot setting while prompt engineering helps only a little. 
CLIP models also perform worse when the question becomes harder (``other'' vs. ``yes/no'').
All these results suggest the need of a deep interactive model and additional pre-training/fine-tuning. 

\begin{table}[]\resizebox{\linewidth}{!}{
\small
\begin{tabular}{lccc}
\toprule
\multicolumn{1}{l}{\multirow{2}{*}{Model}} & \multicolumn{3}{c}{VQA Question Type} \\ \cline{2-4}
\multicolumn{1}{l}{} & \multicolumn{1}{c}{yes/no} & \multicolumn{1}{c}{number} & \multicolumn{1}{c}{other} \\ \toprule
\clipres & 0.037 & 0.057 & 0.0 \\ \midrule
CLIP-ViT-B$_\text{PE}$ & 0.019 & 0.0 & 0.0 \\
CLIP-Res50$_\text{PE}$ & 0.055 & 0.057 & 0.0 \\
CLIP-Res101$_\text{PE}$ & 0.260 & 0.0 & 0.0 \\
CLIP-Res50x4$_\text{PE}$ & 0.446 & 0.118 & 0.034\\
\bottomrule
\end{tabular}}
\vspace{-8pt}
\caption{Zero-shot performance of CLIP on VQA v2.0 \texttt{mini-eval}, ``PE'' denotes we follow similar prompt engineering as suggested in CLIP paper. }
\vspace{-5pt}
\label{tab:zero-shot_vqa}
\end{table}

\noindent\textbf{Unfreezing the Visual Backbone.}
\begin{table}[]\resizebox{\linewidth}{!}{
\begin{tabular}{l|ccc@{}}
\toprule
Feature   & No Pre-train & Pre-train & Diff \\ \midrule
\clipres & 64.66 & 73.92 & +9.26 \\
\clipresbiggest & 69.91 &  76.48 & +6.57 \\
BUTD-Res101  & 66.70 &  72.42 & +5.72 \\
\bottomrule 
\end{tabular}}%
\vspace{-5pt}
\caption{The importance of unfreezing the visual backbone (evaluated on VQA test-dev). CLIP models (unfrozen) get more improvement from V\&L pre-training than BUTD (frozen).}
\vspace{-15pt}
\label{tab:unfreeze}
\end{table}
Because of technical difficulty in fine-tuning the object detector, most \vl models rely on frozen region-based encoders~\cite{lu2019vilbert}.
However, we find that unfreezing the visual backbone (\sref{sec:vl_pretrain}) may bring performance improvement.
Specifically, we test the backbone fine-tuning performance of two CLIP features (i.e., CLIP-Res50, CLIP-Res50x4) on VQA (test-dev) \footnote{For these ablated models without pre-training, we find it beneficial to freeze the visual encoder of our models. We also reduce the batch size to 32 to allow for more gradient updates. Other hyper-parameters are the same as in previous VQA experiments.} and compare with the frozen BUTD-Res101 features. 

Without pre-training, BUTD-Res101 achieves higher performance than \clipres. 
However, after \vl pre-training, \clipres significantly outperforms BUTD-Res101,
because \clipres benefits more from pre-training (+9.25) than BUTD-Res101 (+5.72).
This suggests that unfreezing the visual backbone during pre-training allows \clipres to adapt to the pre-training task. 
We hope that our finding inspires future work to further explore unfreezing the visual backbone in \vl models when computational budget allows.

\begin{figure}[!ht]
    \includegraphics[width=\linewidth]{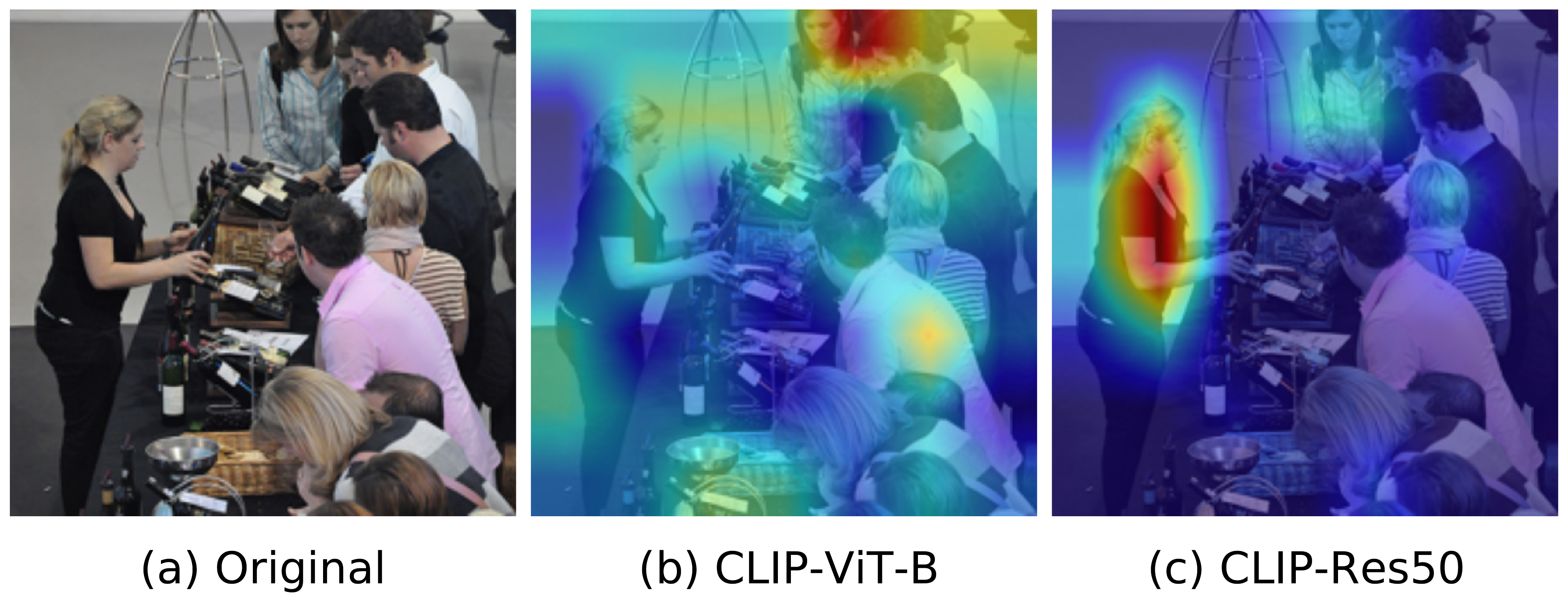}
    \caption{Grad-CAM Visualization of \clipvit and \clipres for the question ``What color is the woman's shirt on the left?''.}
    \label{fig:grad_cam}
\end{figure}

\noindent\textbf{Low Detection Performance of \clipvit.} 
As shown in \tref{tab:vqa_result} and \tref{tab:caption_result}, \clipvit with grid features has a large performance degradation compared with  other models. 
We hypothesize that such decrease is due to the lack of visual localization inside the ViT feature map since different pooling strategies may affect the localization ability~\cite{zhou2016learning}.
We thus follow \newcite{jiang2020defense} to train a detector on Visual Genome over the \clipvit grid feature maps to confirm it.
We find that the Average Precision (AP) of \clipvit is only 0.03, which is much lower than its \imagenetres alternative (3.14 as we reproduced).

\noindent\textbf{Qualitative Comparison of CLIP Variants.}
As we discussed above, we suspect that \clipvit lacks certain localization ability. To understand this better, we perform Gradient-Based Localization (Grad-CAM)~\cite{selvaraju2017grad} to visualize the salient regions of CLIP models. 
The qualitative example in \fref{fig:grad_cam} clearly shows \clipres localizes the sentence ``What  color  is  the woman’s shirt on the left?'' better than \clipvit. 
We provide more qualitative examples in the Appendix.

\label{sec:discussion}

%% file: _s5_discussion.tex

%% file: _s6_conclusion.tex
\section{Conclusions}
\label{sec:conclusions}
In this paper, we propose to leverage CLIP as the visual encoder for different \vl models across various tasks. 
We experiment with two approaches: in the first, we directly plug CLIP in task-specific fine-tuning; 
in the second, we integrate CLIP with \vl pre-training and fine-tune on downstream tasks afterwards. 
A variety of substantial experiments on different \vl tasks demonstrates that \OURS and \OURSP can achieve competitive or better performance as compared to strong baselines.
Analyses from different perspectives explain certain intriguing phenomena and offer new directions for future \vl research. 



%% file: _s7_appendix.tex
\appendix
\section{Appendix}
\label{appendix:Impl}
\subsection{Visual Question Answering}
\paragraph{Model Architecture} Pythia encodes the question with an attention-based GRU~\citep{chung2014empirical} network and fuse the information with a multi-modal factorized bilinear pooling network.
MCAN takes a LSTM~\citep{hochreiter1997long} as question encoder and an encoder-decoder based modular co-attention network for fusing multiple representations.  
Both models employ an output classifier on top of the fused representation to predict the final answer. 
To integrate CLIP for the VQA models, we extract grid features using CLIP. 
For \clipvit models, we reshape the patch representation from the final layer into grid features. 
For CLIP-ResNet models, we simply take the grid features from the last layer before the pooling. 

\paragraph{Implementation Details}
We follow \cite{jiang2020defense} to resize all input images 
to have a maximum shorter side of 600 pixels (longest 1000) when keeping the aspect ratio fixed. 
For training the detector on the VG dataset, we replace the backbone with CLIP visual module using implementation of Faster R-CNN in  Detectron2\footnote{\href{https://github.com/facebookresearch/detectron2}{https://github.com/facebookresearch/detectron2}}. 
For training VQA models, we use hyperparameters of the open-source implementation\footnote{\href{https://github.com/facebookresearch/mmf}{https://github.com/facebookresearch/mmf}} from \cite{jiang2020defense} for the large version of the MCAN and base version of Pythia. 

\subsection{Image Captioning}
\paragraph{Implementation Details}
For training, we follow the `long epoch' hyperparameter of the publicly available implementation~\footnote{\href{https://github.com/ruotianluo/self-critical.pytorch}{https://github.com/ruotianluo/self-critical.pytorch}}. 
During the self-critical stage, we sample 5 captions for each image as in~\newcite{luo2020better}. 
For training objective, we experiment with the Self-Critical Sequence Training (SCST) in~\newcite{rennie2017self}, where CIDEr~\cite{vedantam2015cider} metric is optimized using REINFORCE algorithm~\cite{williams1992simple}.

\subsection{Vision-and-Language Navigation}
\paragraph{Model}
For the model architecture, we experiment with the basic attentive neural agent as in~\newcite{fried2018speaker}.

The agent model (i.e., another LSTM) then attends to the visual features and the language representations to predict the actions.
At each time step $t$, the agent attends to the panoramic views $\{v_{t,i}\}_i$ and the instruction $\{w_j\}$ to make the action.
The panoramic view is processed with a pre-trained visual encoder (e.g., $\mathrm{ResNet}$) and the instructions are processed by a language LSTM~\cite{hochreiter1997long}, denoted $\mathrm{LSTM}_\textsc{l}$.
The agent model, $\mathrm{LSTM}_\textsc{a}$, then attends to the visual features and the language representations to predict the actions.
\begin{align}
    g_{t,i} &= \mathrm{ResNet}(v_{t,i}) \\ 
    x_1, \ldots, x_l &= \mathrm{LSTM}_\textsc{l}(w_1, \ldots, w_l) \\
    \mathit{input}_t &= [\mathrm{Attn}(h_{t-1}\!,\!\{g_{t,i}\})\!,\! \mathrm{Attn}(h_{t-1}\!,\!\{x_j\})] \\ 
    h_{t}, c_{t} &= \mathrm{LSTM}_\textsc{a}(\mathit{input_t}, h_{t-1}, c_{t-1})
\end{align}
where $h_t$ and $c_t$ are the hiddens and states of the action LSTM at time step $t$, respectively.
Please refer to \newcite{fried2018speaker} for the implementation details.

\paragraph{Implementation Details}
We apply our model to two vision-and-language navigation datasets: Room-to-Room (R2R, \newcite{anderson2018vision}) and Room-across-Room (RxR, \newcite{ku2020room}).
R2R is built on the indoor environments from the MatterPort3D dataset~\cite{chang2017matterport3d}.
The environments are split into training (61 environments), unseen validation (11 environments), and unseen test (18 environments).
The agent is trained on the training environments (with 14,025 navigation instructions) and tested on separate sets of environments (2,349 in the unseen-validation and 4,173 in the unseen-test).
RxR extends the R2R dataset with multiple languages and follow the environment split.
Besides the multilingual nature, RxR is also more diverse in the navigation paths and richer in the present language. 
For R2R dataset, we follow the hyperparameter (e.g., batch size, learning rate, optimizer) of the publicly available implementation~\footnote{\href{https://github.com/airsplay/R2R-EnvDrop}{https://github.com/airsplay/R2R-EnvDrop}} R2R-EnvDrop~\cite{tan2019learning} and replace the input features~\footnote{\href{https://github.com/peteanderson80/Matterport3DSimulator}{https://github.com/peteanderson80/Matterport3DSimulator}} with the CLIP features.
To reduce the computational cost, the features are pre-extracted and frozen during the training of the navigational agent.
For RxR dataset, we take the processed multilingual data provided in \newcite{li2021improving} with Stanza tokenizers~\cite{qi2020stanza}.
Since RxR dataset contains instructions longer than R2R, we change the maximum input length to $160$ (from $80$) and increase the imitation learning ratio from $0.2$ to $0.4$ to stabilize the training. 
Other training hyperparameters of RxR are the same as R2R.
The models are trained on one RTX 2080 Ti GPU. 
It takes 1 days to converge in R2R and about 1.5 days to converge in RxR.
We report two significant digits for R2R unseen test results following the leaderboard convention.

\paragraph{Results Comparison to Grid Features}
\begin{table}[]
\begin{tabular}{@{}llcc@{}}
\toprule
Feature         & Dimension & SR   & \textbf{SPL}  \\ \midrule
ImageNet-Res152 & 2048      & 48.2 & 44.4 \\ 
CLIP-Res50      & 1024      & 52.6 & 47.4 \\ \midrule
Grid-Res50      & 2048      & 47.6 & 44.7 \\
Grid-ResX101    & 2048      & 46.5 & 43.2 \\
Grid-ResX152    & 2048      & 47.8 & 44.6 \\ \bottomrule
\end{tabular}%
\caption{Comparison between grid features, CLIP features, and ImageNet-trained features on the R2R dataset. `SR' and `SPL' are success rate and success rate weighted by path length. }
\label{tab:nav_grid}
\end{table}

In the main paper, we compare the results regarding the ImageNet-pre-trained ResNet-152. 
We also report the comparison to grid features~\newcite{jiang2020defense} that is trained with detection dataset.
\newcite{jiang2020defense} showed that the results with these features are comparable to the original bottom-up attention with a heavy detection module.
The same as the VQA task in Section 3.1, we test the performance of these detection-trained grid features on VLN tasks.
Specifically, we use the mean pooling of the feature map as the representation of each view following previous works~\cite{anderson2018vision}.
As shown in Table~\ref{tab:nav_grid}, under the same ResNet50 backbone~\footnote{The CLIP model uses an attention pooling module and makes modifications over the original ResNet~\cite{he2016deep} backbone.}, we find that the detection-trained grid features are on par with the classification-trained grid features, still showing a gap to the contrastive-trained grid features.
We hypothesize that the grid features inject regional knowledge into the dense feature map thus showing good results with grid-based modules (as shown in Section~\ref{sec:vqa}). 
However, pooling the feature map into a single feature vector (as in previous VLN works) leads to a loss of this dense information.

\subsection{Details of \OURSP}

\paragraph{Pre-training} We pre-train with a batch size of 512. 
The Transformer is initialized from \bertbase and optimized with an AdamW~\cite{loshchilov2017decoupled} optimizer. 
We use a linearly-decaying schedule and a peak learning rate of $1\times10^{-4}$ for the model with \clipres and $5\times10^{-5}$ for the model with \clipresbiggest. 
The ResNet is initialized from CLIP and we use SGD with a learning rate of $3\times10^{-3}$. We decay the learning rate of SGD at epochs 12, 17 by a factor of 10. 
Per the suggestion of \citet{tan2019lxmert}, we only add the visual question answering loss during the later stage of the pre-training (the last 11 epochs) as the model is prone to overfit to the visual question answering loss. 
The model is trained on 8 Nvidia A100 GPUs and the pre-training takes around 5 days.

\begin{figure}[!ht]
    \includegraphics[width=\linewidth]{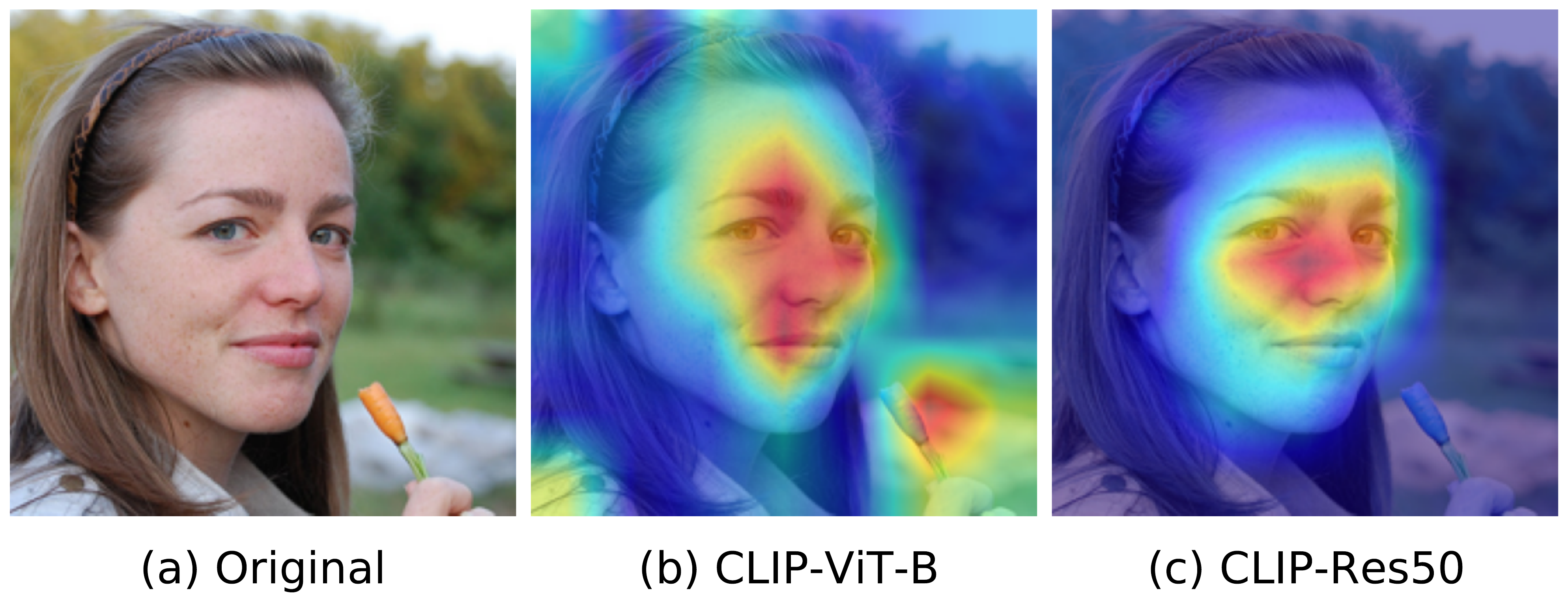}
    \caption{Grad-CAM Visualization of \clipvit and \clipres for the question ``What color are her eyes?''.}
    \label{fig:grad_cam_2}
\end{figure}

\begin{figure}[!ht]
    \includegraphics[width=\linewidth]{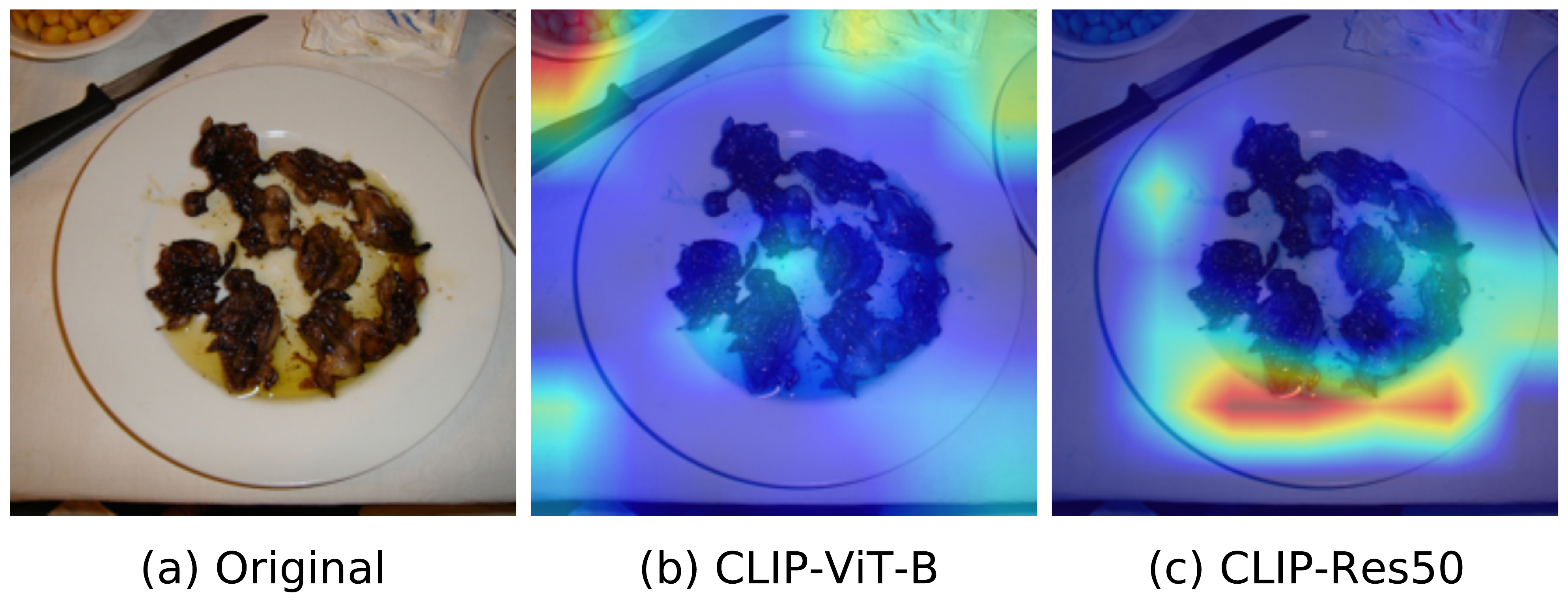}
    \caption{Grad-CAM Visualization of \clipvit and \clipres for the question ``What is just above the plate?''.}
    \label{fig:grad_cam_3}
\end{figure}

\paragraph{Fine-tuning}
We fine-tune \OURSP on three tasks: VQA v2.0, SNLI-VE, and GQA. We introduce the task specifics and fine-tuning hyper-parameters in the following.

Every example in VQA consists of an image and a question, where the task is to predict the correct answer.  We use the Karpathy split for training and validation \citep{karpathy2015deep}. We  fine-tune the model with the binary cross-entropy loss for 5 epoch with a batch size of 256.  The Transformer is optimized with AdamW and a peak learning rate of $5\times10^{-5}$.  The ResNet is optimized with SGD and an initial learning rate of $1\times10^{-3}$. We decay the learning rate of ResNet by a factor of 10 after epoch 3.

SNLI-VE is a three-way classification task, which involves determining the relation between an image and a sentence. 
The three possible relations include entailment, contradiction, and neutral. 
We fine-tune the model with the negative log-likelihood loss for 2 epoch with a batch size of 256. 
The Transformer is optimized with AdamW and a peak learning rate of $5\times10^{-5}$. 
The ResNet is optimized with SGD and an initial learning rate of $1\times10^{-3}$. We decay the learning rate of ResNet by a factor of 10 after epoch 1. 

GQA follows the format of VQA but the questions and answers of GQA are automatically generated from ground-truth scene graphs. We use the same hyper-parameters as in VQA.

\subsection{More Qualitative Examples}
Here we present more qualitative examples using (Grad-CAM)~\cite{selvaraju2017grad} to visualize the salient regions of CLIP models. 
\fref{fig:grad_cam_2} and \fref{fig:grad_cam_3} suggest that \clipres localizes the sentence better than \clipvit.